\documentclass[a4paper,fleqn]{cas-dc}

\usepackage[numbers]{natbib}
\usepackage{graphicx}
\usepackage{booktabs}
\usepackage{subcaption}
\usepackage{amsmath}
\usepackage{amsthm}
\usepackage{array}
\usepackage{amssymb}
\usepackage{pifont}
\usepackage{algorithm}
\usepackage{graphicx}
\usepackage{algpseudocode}
\usepackage{caption}
\usepackage{float}
\usepackage{amsmath,amssymb,amsfonts}
\usepackage{xcolor}
\usepackage{amsmath,amssymb,mathtools}

\usepackage{pifont}               

\usepackage{orcidlink}
\usepackage{pifont}
\usepackage{geometry}
\usepackage{adjustbox}
\usepackage{rotating}
\usepackage{comment}
\usepackage{subcaption}

\hypersetup{
    colorlinks=true,
    linkcolor=blue,
    citecolor=blue,
    filecolor=blue,
    urlcolor=blue
}

\def\tsc#1{\csdef{#1}{\textsc{\lowercase{#1}}\xspace}}
\tsc{WGM}
\tsc{QE}

\begin{document}
\let\WriteBookmarks\relax
\def\floatpagepagefraction{1}
\def\textpagefraction{.001}

\shorttitle{CrosGrpsABS: Cross-Attention over Syntactic and Semantic Graphs for Aspect-Based Sentiment Analysis in a Low-Resource Language
}    

\shortauthors{}  

\title [mode = title]{CrosGrpsABS: Cross-Attention over Syntactic and Semantic Graphs for Aspect-Based Sentiment Analysis in a Low-Resource Language}  


%

\author[1]{Md. Mithun Hossain}[orcid=0009-0001-4883-1802]

\ead{mithunhossain@bubt.edu.bd}

\credit{Conceptualization of this study, Methodology, Data Curation, Software, Writing - Original Draft}

\author[1]{Md. Shakil Hossain}[orcid=0009-0009-1584-3282]
\ead{shakilhossain@bubt.edu.bd}

\credit{Investigation, Methodology, Formal analysis, Visualization, Writing - Original Draft}

\author[1]{Sudipto Chaki}[orcid=0000-0002-7286-6722] 
\ead{sudipto@bubt.edu.bd}

\credit{Validation, Supervision, Writing - Original Draft, Writing - Review \& Editing}

\author[2]{Md. Rajib Hossain}[orcid=0000-0002-7941-9124]
\ead{rajcsecuetphd@cuet.ac.bd}

\credit{Validation, Supervision, Writing - Original Draft, Writing - Review \& Editing}

\affiliation[1]{organization={Department of Computer Science and Engineering, Bangladesh University of Business and Technology},
            addressline={Mirpur-2}, 
            city={Dhaka},
            postcode={1216}, 
            country={Bangladesh}}

\affiliation[2]{organization={Department of Computer Science and Engineering, Chittagong University of Engineering and Technology},
            addressline={Raozan}, 
            city={Chittagong},
            postcode={4349}, 
            country={Bangladesh}}







\cortext[1]{Corresponding author: S. Chaki (sudipto@bubt.edu.bd)}



\begin{abstract}  
Aspect-Based Sentiment Analysis (ABSA) is a fundamental task in natural language processing, offering fine-grained insights into opinions expressed in text. While existing research has largely focused on resource-rich languages like English which leveraging large annotated datasets, pre-trained models, and language-specific tools. These resources are often unavailable for low-resource languages such as Bengali. The ABSA task in Bengali remains poorly explored and is further complicated by its unique linguistic characteristics and a lack of annotated data, pre-trained models, and optimized hyperparameters. To address these challenges, this research propose \textit{CrosGrpsABS}, a novel hybrid framework that leverages bidirectional cross-attention between syntactic and semantic graphs to enhance aspect-level sentiment classification. The \textit{CrosGrpsABS} combines transformer-based contextual embeddings with graph convolutional networks, built upon rule-based syntactic dependency parsing and semantic similarity computations. By employing bidirectional cross-attention, the model effectively fuses local syntactic structure with global semantic context, resulting in improved sentiment classification performance across both low- and high-resource settings. We evaluate  \textit{CrosGrpsABS} on four low-resource Bengali ABSA datasets and the high-resource English SemEval 2014 Task 4 dataset. The \textit{CrosGrpsABS} consistently outperforms existing approaches, achieving notable improvements, including a 0.93\% F1-score increase for the Restaurant domain and a 1.06\% gain for the Laptop domain in the SemEval 2014 Task 4 benchmark.  
\end{abstract}


\maketitle

\section{Introduction}
Aspect-based Sentiment Analysis (ABSA) has attracted considerable attention in the field of Natural Language Processing (NLP), as it enables fine-grained sentiment detection by identifying sentiment polarities associated with specific aspects or attributes \cite{pontiki2014semeval, zhang2019aspect, an2022aspect}. Unlike traditional sentiment analysis, which typically assesses sentiment at the document or sentence level, ABSA provides more detailed and actionable insights by focusing on individual features or components~\cite{liu2012sentiment}. In the era of the Internet and widespread engagement on social media platforms, ABSA has become an essential tool for analyzing product reviews, public opinions, and user feedback, playing a vital role in enhancing the reliability and trustworthiness of online services and recommendations. While many useful tools and applications have been developed ABSA in high-resource languages, there remains a significant lack of accessible tools and research resources for low-resource languages such as Bengali.

Although Bengali is the seventh most spoken language globally, with approximately 273 million native speakers and more than 39 million second-language speakers worldwide~\cite{hossain2025authornet}, it remains notably underrepresented in the context of fine-grained sentiment analysis. Existing research on Bengali sentiment analysis focuses primarily on broader classification tasks, such as document-level or sentence-level sentiment detection~\cite{banik2019survey, su2024enhanced}. For example, Banik et al.~\cite{banik2019survey} provided an overview of Bengali sentiment analysis trends, but their findings suggest that aspect-based approaches have been explored to a limited extent. This gap is further exacerbated by unique linguistic complexities in Bengali, which include rich morphology, diverse syntactic structures, and frequent implicit sentiment expressions. Moreover, the limited availability of large-scale, high-quality annotated datasets, along with the scarcity of accurate Bengali-specific dependency parsers and part-of-speech (POS) taggers, poses substantial challenges in developing robust ABSA systems for this language.

Recent advances leveraging transformer-based models such as BERT~\cite{devlin2019bert}, RoBERTa~\cite{liu2019roberta}, and graph-based neural network architectures~\cite{zhang2019aspect, wang2022survey} have significantly propelled ABSA research, particularly for high-resource languages such as English. However, other studies indicate that these models rely heavily on substantial annotated data and language-specific resources, creating significant barriers for low-resource language applications~\cite{wang2022survey, nazir2020survey}.

These challenges lead to several key research questions (RQs):  \textbf{(RQ1)}~How do syntactic dependency and semantic similarity graphs—individually and in combination—affect the performance and interpretability of aspect-based sentiment analysis models?  
\textbf{(RQ2)}~How can existing transformer-based and graph-based ABSA techniques be effectively adapted across languages with varying resource availability, such as low-resource languages like Bengali and high-resource languages like English?

\begin{figure}[htbp]
  \centering
  \includegraphics[scale=0.65]{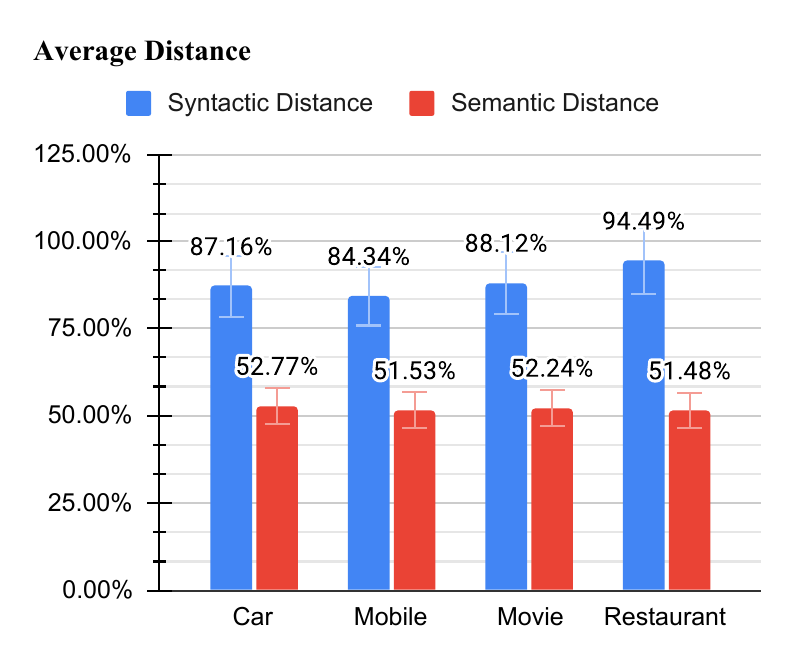}
  \caption{\small Comparison of syntactic and semantic distances across different aspects (Car, Mobile, Movie, and Restaurant). The results indicate that syntactic distances (blue) are generally larger than semantic distances (red), highlighting the need for a mechanism that effectively integrates syntactic and semantic information for aspect-based sentiment analysis in Bengali texts.}
  \label{avg_distance}
\end{figure}

Syntactic distance measures the structural difference between words in a sentence based on grammatical rules, while semantic distance assesses the conceptual similarity between words using their contextual meanings. In the context of aspect-based sentiment analysis, these distances help in understanding how words relate to one another from both a structural and a meaning perspective. \textcolor{blue}{Figure \ref{avg_distance}} shows the average syntactic and semantic distances for four aspects: Car, Mobile, Movie, and Restaurant. Syntactic distances, represented in blue, tend to have higher values indicating more variability in grammatical relationships, whereas semantic distances, shown in red, are generally lower, reflecting more consistent contextual connections. This comparison underscores the importance of integrating both types of information to capture the full spectrum of linguistic nuances in sentiment analysis.

To address the research questions, we propose a novel framework, \textbf{CrosGrpsABS}, that unifies syntactic and semantic graph representations through a bidirectional cross-attention mechanism. In response to \textbf{RQ1}, the framework investigates how syntactic dependency and semantic similarity graphs—individually and jointly—contribute to both the performance and interpretability of ABSA models. For \textbf{RQ2}, it explores how such a hybrid model can be effectively adapted across languages with varying resource availability, particularly focusing on Bengali as a low-resource language and English as a high-resource benchmark.
\textbf{CrosGrpsABS} constructs two complementary graphs: a syntactic graph that captures local grammatical relationships using dependency parsing tailored to the linguistic characteristics of Bengali, and a semantic graph that captures global contextual relationships by computing cosine similarities between transformer-based word embeddings. These graphs are dynamically fused through a bidirectional cross-attention mechanism that aligns and refines node representations based on cross-graph signals—allowing syntactic and semantic cues to reinforce each other. This design not only enhances sentiment prediction but also offers interpretability by highlighting influential syntactic structures and semantic associations.

The key contributions and possible answers to the research questions (ARQs) related to the proposed \textbf{CrosGrpsABS} model are summarized as follows:

\begin{itemize}
    \item \textbf{ARQ1:} We present a unified graph construction strategy that captures both syntactic dependency and semantic similarity information using transformer-based contextual embeddings. These graphs are integrated through a novel bidirectional cross-attention mechanism, allowing the model to effectively leverage both syntactic structure and semantic context to improve the interpretability and performance of aspect-based sentiment analysis.

    \item \textbf{ARQ2:} We propose \textbf{CrosGrpsABS}, a hybrid ABSA framework that combines transformer-based and graph-based techniques, and demonstrate its adaptability to languages with varying resource availability. Through extensive experiments on a low-resource Bengali ABSA dataset and high-resource English datasets (SemEval 2014 Task 4: Restaurant and Laptop), we show that our model generalizes well across languages, highlighting its effectiveness in both constrained and rich-resource settings.
\end{itemize}

The remainder of this paper is organized as follows. \textcolor{blue}{Section}~\ref{literature} reviews the related work on ABSA, graph-based models, and sentiment analysis for low-resource languages. \textcolor{blue}{Section}~\ref{banglabca} presents the architecture of the proposed \textbf{CrosGrpsABS}. \textcolor{blue}{Section}~\ref{experimental} describes the experimental setup, including datasets and evaluation metrics. \textcolor{blue}{Section}~\ref{resullts} discusses the results and compares them with existing baseline models. Finally, \textcolor{blue}{Section}~\ref{conclusion} concludes the study and suggests future research directions.

\section{Literature Review}
\label{literature}

\textbf{Aspect-Based Sentiment Analysis in High-Resource Languages:}  
Aspect-Based Sentiment Analysis (ABSA) has witnessed substantial progress in high-resource languages, particularly English, owing to the availability of large-scale annotated datasets, advanced language tools, and powerful pre-trained language models. Transformer-based architectures such as BERT~\cite{joloudari2023bert}, RoBERTa~\cite{liao2021improved}, and their specialized ABSA variants have significantly improved sentiment classification performance by capturing complex contextual dependencies through self-attention mechanisms~\cite{letarte2018importance}. These models often serve as the backbone for sentence-level and aspect-level classification tasks.  
Beyond sequential models, graph-based approaches have gained prominence for their ability to incorporate syntactic and semantic structures explicitly \cite{su2024enhanced}. Dependency tree-based models~\cite{di2013sentiment, zhang2023enhanced} and graph neural networks (GNNs) such as GCNs and TextGCNs~\cite{liang2022aspect, yin2024textgt} have been employed to construct graph representations of sentence structure, enabling the learning of relational patterns between aspect and opinion terms. Semantic graphs based on similarity measures~\cite{li2021dual} further enhance global contextual understanding \cite{nazir2022iaf, qiu2023modeling}. However, many existing models in high-resource contexts treat syntactic and semantic features independently, limiting their capacity to leverage their complementary strengths for interpretability and performance.  

\textbf{Aspect-Based Sentiment Analysis in Low-Resource Languages (e.g., Bengali):}  
Despite the global prevalence of languages like Bengali, ABSA research in low-resource settings remains underdeveloped. Bengali-specific sentiment analysis efforts have predominantly addressed coarse-grained tasks at the document or sentence level~\cite{banik2019survey, hossain2022sentence}, with limited exploration of fine-grained ABSA. The scarcity of annotated aspect-level datasets, lack of robust dependency parsers, and the unavailability of optimized Bengali-specific pre-trained models constrain the development of syntactic and semantic representations~\cite{nazir2020issues}. Furthermore, Bengali's complex morphological structures and syntactic flexibility introduce challenges not observed in high-resource languages.  
Recent studies have attempted to adapt multilingual transformers (e.g., mBERT, XLM-R) for Bengali ABSA~\cite{meetei2021low, roy2023review}, but performance often suffers due to mismatches in linguistic structure and domain specificity. Graph-based methods are even less explored in Bengali due to the difficulty of building accurate syntactic trees and semantic graphs without language-specific tools. This underscores the need for an adaptable ABSA framework that can overcome these limitations by integrating syntactic and semantic structures.

\textbf{Research Gaps and Mitigation:}  
Although both syntactic and semantic cues are vital for ABSA, most existing works treat them as disjoint components. Some recent research has attempted to bridge this gap using attention or fusion mechanisms, but few offer a unified and interpretable integration. Models that employ both syntactic dependencies and semantic similarities often fail to capture cross-level interactions in a structured and bidirectional manner~\cite{li2021dual, li2022dualgcn}. Furthermore, many of these models lack explainability, making it difficult to interpret how graph structures influence prediction outcomes.  
The problem becomes more pronounced in low-resource settings, where the challenges of representation construction are magnified. The absence of unified graph fusion strategies, especially for low-resource languages like Bengali, limits the development of generalized ABSA models. Moreover, cross-lingual adaptability has rarely been explored through graph-based techniques, which further restricts their application in multilingual contexts.  
Thus, a clear research gap exists in designing an interpretable graph-based ABSA framework that can jointly leverage syntactic structure and semantic context while maintaining adaptability across languages. This motivates the development of our proposed \textbf{CrosGrpsABS} model (i.e., \textcolor{blue}{Figure} \ref{overview}), which unifies syntactic (i.e., \textcolor{blue}{Section} \ref{syantactic}) and semantic (i.e., \textcolor{blue}{Section} \ref{semantic}) graphs using a novel bidirectional cross-attention mechanism (i.e., \textcolor{blue}{Section} \ref{cross-attention}, and \textcolor{blue}{Figure} \ref{attention}) to address both \textbf{RQ1} (graph interaction shown in \textcolor{blue}{Figure} \ref{gnn_layers_car} and interpretability shown in \textcolor{blue}{Figure} \ref{token}) and \textbf{RQ2} (cross-lingual adaptability), showing promising results (i.e., \textcolor{blue}{Section} \ref{resullts}) in both Bengali (i.e., \textcolor{blue}{Table} \ref{model_comparison}, \ref{ablation_study}, and \ref{case}) and English ABSA benchmarks (i.e., \textcolor{blue}{Table} \ref{multirow_example}).

\section{Proposed Methodology}\label{banglabca}
\begin{figure*}[h]
    \centering
    \includegraphics[scale=0.2]{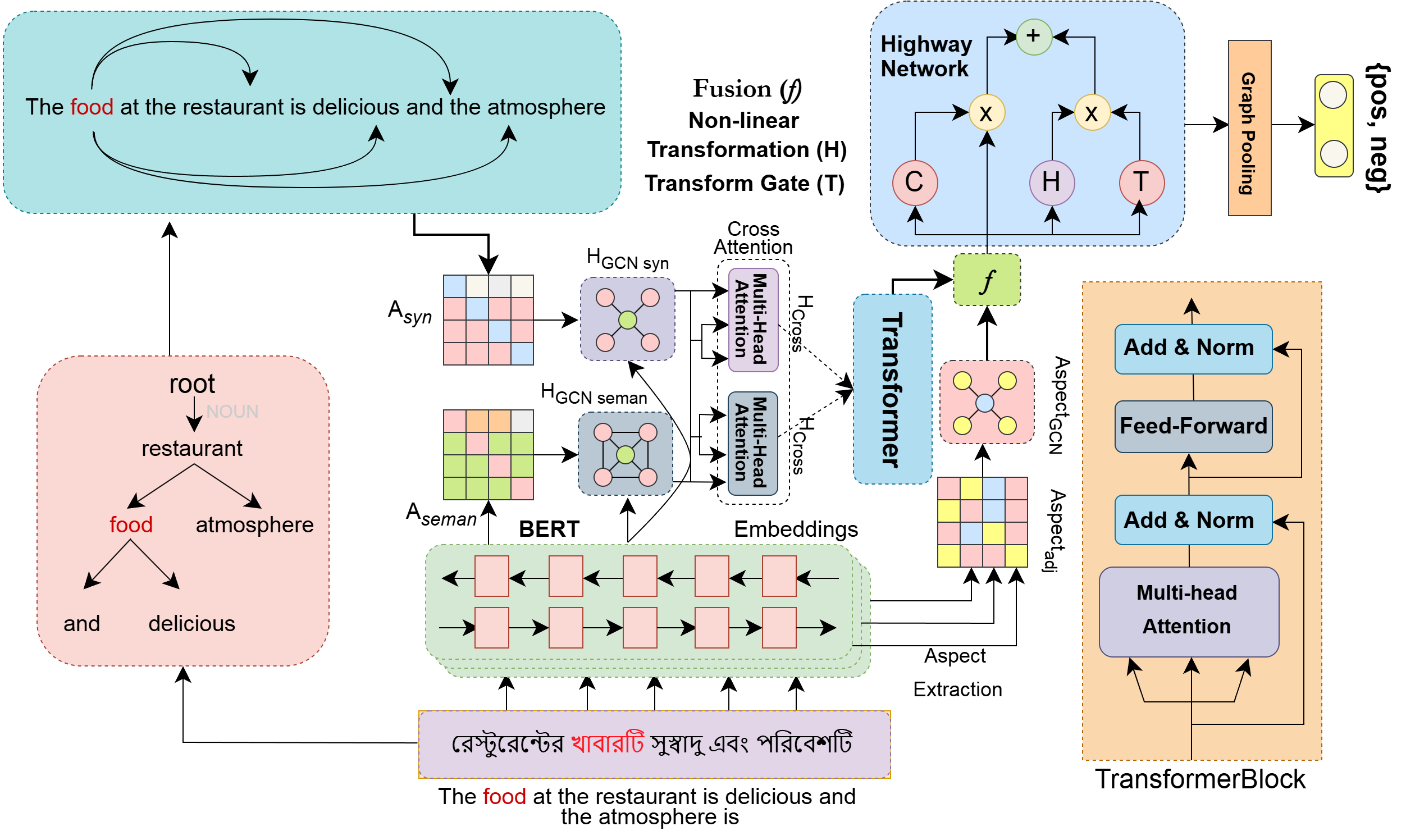}
    \caption{\small The overall \textbf{CrosGrpsABS} system flow, depicting the construction of syntactic and semantic graphs, cross-attention fusion with transformer embeddings, and final sentiment classification using a highway gating strategy.}
    \label{overview}
\end{figure*}

Aspect-based sentiment analysis (ABSA) aims to determine the sentiment polarity expressed toward specific aspects within a given sentence. Formally, consider a sentence \( s = \{w_1, w_2, \ldots, w_n\} \) and an aspect \( a = \{a_1, a_2, \ldots, a_m\} \subset s\). Here, \(m\) denotes the end position of the aspect in \(s\). When a sentence contains multiple aspects, each aspect is treated as a separate instance for sentiment classification. The goal is to predict the sentiment polarity (e.g., positive, negative, or neutral) associated with each aspect \(a\). \textcolor{blue}{Figure~\ref{overview}} illustrates the overall architecture of \textbf{CrosGrpsABS}, which addresses the aforementioned problem setting. Given a sentence-aspect pair \((s, a)\), each token in the sentence is first mapped to its corresponding embedding through an embedding table \( \mathbf{E} \in \mathbb{R}^{|V| \times d_e}, \) where \(|V|\) is the vocabulary size and \(d_e\) is the embedding dimension. We concatenate the sentence and aspect to form a single input sequence for a transformer encoder (e.g., BERT~\cite{devlin2019bert} or XLM-RoBERTa \cite{liu2019roberta}), formatting it as:
\(\text{Input} = [\text{[CLS]}] \; w_1,\ldots,w_n \; [\text{SEP}] \; a_1,\ldots,a_m \; [\text{SEP}],
\) where \texttt{[CLS]} serves as a classification token and \texttt{[SEP]} tokens separate the sentence and aspect. Passing this sequence through the transformer yields contextual embeddings
\(
\mathbf{H} = [\mathbf{h}_1,\mathbf{h}_2,\ldots,\mathbf{h}_T] \in \mathbb{R}^{T \times d},
\) with \(T = n + m + 3\) accounting for the added special tokens, and \(d\) indicating the hidden dimensionality.

These contextual representations, \(\mathbf{H}\), provide initial node features for two distinct graphs: a \emph{syntactic graph}, constructed through rule-based dependency parsing to capture local grammatical structure, and a \emph{semantic graph}, built using cosine similarity of transformer-based embeddings to capture global contextual relationships. The subsequent modules in \textbf{CrosGrpsABS} leverage these graphs through a bidirectional cross-attention mechanism, thus integrating syntactic and semantic information to more accurately predict aspect-level sentiment.

\subsection{Syntactic Graph}
\label{syantactic}

To encode syntactic relationships in our model, we begin by applying a rule-based parser (or a dependency parser) to each sentence, identifying part-of-speech (POS) tags and any dependency-like links between tokens \cite{zhang2019aspect}. The objective is to determine, for each token \(i\), which tokens \(j\) are syntactically related. Once these relationships are identified, we construct the adjacency matrix \(\mathbf{A}_{\text{syntax}}\) by iterating through tokens and marking pairs \((i, j)\) that are identified as valid dependencies (for example, subject, object or modifier links). Formally, we define \(\mathbf{A}_{\text{syntax}}\) as shown in \textcolor{blue}{Equation~\eqref{eq:Asyntax}}, where:
\begin{equation}
\mathbf{A}_{\text{syntax}}[i,j] =
\begin{cases}
1, & \text{if } j \text{ is syntactically related to } i, \\
0, & \text{otherwise}.
\end{cases}
\label{eq:Asyntax}
\end{equation}
This binary setup captures whether a direct syntactic link exists between two tokens. In practie, fractional or real-valued weights can be assigned in \(\mathbf{A}_{\text{syntax}}\) rather than a strict 0/1 approach. For example, edges can be scaled based on dependency type (e.g., subject vs.\ object) or the distance between tokens. By explicitly modeling syntactic structure in this manner, the model can better interpret each token’s grammatical function relative to the aspect. Identifying a subject-object relationship, for instance, can reveal whether the aspect is performing an action (potentially implying a certain polarity) or is affected by another entity in the sentence.

\subsection{Semantic Similarity Graph}
\label{semantic}

We construct a semantic similarity graph to capture correlations between tokens that share related meanings \cite{huang2019syntax,li2018transformation,zhang2019aspect}. Each token \(w_i\) in the sentence is mapped to a contextual embedding \(\mathbf{h}_i\) derived from a transformer encoder, yielding \(\{\mathbf{h}_1, \mathbf{h}_2, \ldots, \mathbf{h}_n\}\). We then measure the similarity between each pair of tokens \((i, j)\) using the cosine similarity formula given in \textcolor{blue}{Equation~\eqref{eq:semantic_adj}}:
\begin{equation}
\mathbf{A}_{\text{semantic}}[i,j] \;=\; 
\frac{\mathbf{h}_i \cdot \mathbf{h}_j}
     {\|\mathbf{h}_i\|_2 \;\|\mathbf{h}_j\|_2},
\label{eq:semantic_adj}
\end{equation}
which quantifies the degree of semantic relatedness between \(\mathbf{h}_i\) and \(\mathbf{h}_j\). By linking words that share semantic fields or appear in similar contexts, the adjacency matrix \(\mathbf{A}_{\text{semantic}}\) captures long-range or indirect relationships, an essential feature for ABSA where sentiment cues may span distant tokens. To manage graph density, each row can be normalized or thresholded, enabling the model to focus on the most salient semantic connections. Consequently, the semantic similarity graph complements the syntactic graph by emphasizing shared meaning rather than purely structural dependencies.

\subsection{Graph Attention Networks (GAT)}
We denote the initial node features as the transformer outputs, \(\mathbf{H}\). A GAT layer operates over an adjacency matrix \(\mathbf{A}\) (e.g., syntactic or semantic) and updates each node’s representation by attending to its neighboring nodes \cite{velickovic2018graph,kipf2017semi}. The first step in this process is computing the attention coefficients between node \(i\) and its neighbor \(j\):
\begin{equation}
\alpha_{ij} = \text{softmax}_j\Big(\text{LeakyReLU}\big(\mathbf{a}^T [\mathbf{W}\mathbf{h}_i \,\|\, \mathbf{W}\mathbf{h}_j]\big)\Big),
\label{eq:attention}
\end{equation}
where \(\mathbf{W}\) is a learnable weight matrix that projects node features into an attention space, and \(\mathbf{a}\) is a learnable attention vector. The concatenation \([\mathbf{W}\mathbf{h}_i \,\|\, \mathbf{W}\mathbf{h}_j]\) allows the attention mechanism to model interactions between the node and its neighbors, as shown in \textcolor{blue}{Equation~\eqref{eq:attention}}. A LeakyReLU activation function is applied before computing the softmax normalization across all neighbors.

Once the attention coefficients \(\alpha_{ij}\) are computed, they are used to aggregate the information from the neighboring nodes. The updated feature representation for node \(i\) is given by:
\begin{equation}
\tilde{\mathbf{h}}_i = \sigma\Big(\sum_{j \in \mathcal{N}(i)} \alpha_{ij}\, \mathbf{W}\mathbf{h}_j\Big),
\label{eq:update}
\end{equation}
where \(\mathcal{N}(i)\) denotes the set of neighbors for node \(i\), and \(\sigma\) represents a non-linear activation function such as ReLU. This update process ensures that the representation of node \(i\) incorporates important contextual information from its neighbors while filtering out less relevant interactions \cite{zhang2019aspect}.

In the final step of the GAT layer, all updated node representations \(\tilde{\mathbf{h}}_i\) are combined to form the output representation of the entire graph:
\begin{equation}
\mathbf{Z} = [\tilde{\mathbf{h}}_1, \tilde{\mathbf{h}}_2, \ldots, \tilde{\mathbf{h}}_T],
\label{eq:final_z}
\end{equation}
where \(T\) represents the total number of nodes. \textcolor{blue}{Equation~\eqref{eq:final_z}} ensures that all nodes have been updated using attention-based aggregation.

We apply GAT separately to the syntactic and semantic graphs, which results in two different node representations:
\begin{align}
\mathbf{H}_{\text{syntax}} &= \text{GATConv}(\mathbf{H}, \mathbf{A}_{\text{syntax}}), \label{eq:syntax_gat}\\[1mm]
\mathbf{H}_{\text{semantic}} &= \text{GATConv}(\mathbf{H}, \mathbf{A}_{\text{semantic}}). \label{eq:semantic_gat}
\end{align}
Here, \textcolor{blue}{Equation~\eqref{eq:syntax_gat}} represents the GAT operation over the syntactic graph, allowing each token to focus on crucial grammatical relations, while \textcolor{blue}{Equation~\eqref{eq:semantic_gat}} describes the GAT operation over the semantic graph, capturing similarity-based relationships \cite{velickovic2018graph}.

By leveraging GAT, each token dynamically attends to its most relevant neighbors rather than treating all connected nodes equally. The attention mechanism in \textcolor{blue}{Equation~\eqref{eq:attention}} allows the model to assign higher weights to influential words, while \textcolor{blue}{Equation~\eqref{eq:update}} ensures that only meaningful information is aggregated. Furthermore, stacking multiple GAT layers enables the model to capture multi-hop dependencies, which is essential for processing complex Bengali sentences. Ultimately, the combination of transformer-based self-attention with graph-based neighbor attention results in a richer and more structured representation of the input text.

\subsection{Bidirectional Cross-Attention} 
\label{cross-attention}
To effectively merge the graph-based features with the original transformer context, we employ a bidirectional cross-attention mechanism \cite{vaswani2017attention,devlin2019bert}. This mechanism allows the model to compare and fuse information from the transformer embeddings \(\mathbf{H}\) with the GAT embeddings derived from both the syntactic and semantic graphs \cite{hu2020explicit}.

First, for the syntactic branch, we project the transformer embeddings and the syntactic GAT embeddings \(\mathbf{H}_{\text{syntax}}\) into a common attention space. This is achieved by computing the query, key, and value matrices as follows:

\begin{subequations}
\label{eq:syn_group} 
\renewcommand{\theequation}{6\alph{equation}} 
\begin{align}
\mathbf{Q}_{\text{syn}} &= \mathbf{W}_Q^{(\text{syn})}\,\mathbf{H}, 
\label{eq:syn_q}\\
\mathbf{K}_{\text{syn}} &= \mathbf{W}_K^{(\text{syn})}\,\mathbf{H}_{\text{syntax}}, 
\label{eq:syn_k}\\
\mathbf{V}_{\text{syn}} &= \mathbf{W}_V^{(\text{syn})}\,\mathbf{H}_{\text{syntax}},
\label{eq:syn_v}\\[6pt]
\mathbf{C}_{\text{syn}} &= \text{softmax}\!\Bigg(
    \frac{\mathbf{Q}_{\text{syn}}\,\mathbf{K}_{\text{syn}}^T}{\sqrt{d}}
\Bigg)\,\mathbf{V}_{\text{syn}},
\label{eq:syn_attention}
\end{align}
\end{subequations}

In \textcolor{blue}{Equations~(\ref{eq:syn_q})}--\textcolor{blue}{(\ref{eq:syn_v})}, the transformer and syntactic features are mapped into query, key, and value spaces, respectively, while \textcolor{blue}{Equation~(\ref{eq:syn_attention})} computes the syntax-based attention output \(\mathbf{C}_{\text{syn}}\). The division by \(\sqrt{d}\) (where \(d\) is the dimensionality of the projection space) helps stabilize gradients \cite{vaswani2017attention}.

Analogously, for the semantic branch, we compute the corresponding query, key, and value matrices using the semantic GAT embeddings \(\mathbf{H}_{\text{semantic}}\):

\begin{subequations}
\label{eq:sem_group}  
\renewcommand{\theequation}{7\alph{equation}} 
\begin{align}
\mathbf{Q}_{\text{sem}} &= \mathbf{W}_Q^{(\text{sem})}\,\mathbf{H}, 
\label{eq:sem_q}\\
\mathbf{K}_{\text{sem}} &= \mathbf{W}_K^{(\text{sem})}\,\mathbf{H}_{\text{semantic}}, 
\label{eq:sem_k}\\
\mathbf{V}_{\text{sem}} &= \mathbf{W}_V^{(\text{sem})}\,\mathbf{H}_{\text{semantic}},
\label{eq:sem_v}\\[6pt]
\mathbf{C}_{\text{sem}} &= \text{softmax}\!\Bigg(
  \frac{\mathbf{Q}_{\text{sem}}\,\mathbf{K}_{\text{sem}}^T}{\sqrt{d}}
\Bigg)\,\mathbf{V}_{\text{sem}},
\label{eq:sem_attention}
\end{align}
\end{subequations}

Here, \textcolor{blue}{Equations~(\ref{eq:sem_q})}--\textcolor{blue}{(\ref{eq:sem_v})} map the transformer and semantic features into query, key, and value spaces, while \textcolor{blue}{Equation~(\ref{eq:sem_attention})} yields the semantic attention output \(\mathbf{C}_{\text{sem}}\). 

Finally, the outputs from both the syntactic and semantic cross-attention blocks are concatenated to form a unified representation:
\begin{equation}
\mathbf{H}_{\text{cat}} = [\mathbf{C}_{\text{syn}} \,\|\, \mathbf{C}_{\text{sem}}] \in \mathbb{R}^{T \times 2d}.
\label{eq:concat}
\end{equation}
\textcolor{blue}{Equation~(\ref{eq:concat})} provides the fused context \(\mathbf{H}_{\text{cat}}\), which contains enriched information from both structural (syntactic) and semantic perspectives.

This bidirectional cross-attention mechanism enables the transformer embeddings to interact closely with graph-based representations, allowing for a dynamic fusion where local structural signals (e.g., dependency edges) and global semantic patterns jointly influence the final feature representation \cite{devlin2019bert}. Such synergy is crucial for aspect-based sentiment analysis (ABSA), where an aspect’s sentiment depends on both detailed local context and the overall semantic coherence of the sentence \cite{hu2020explicit}.

\subsection{Transformer Encoder Refinement}

To further integrate the syntax- and semantic-enriched features, we feed \(\mathbf{H}_{\text{cat}}\) into a lightweight Transformer encoder as shown in \textcolor{blue}{Equation~\eqref{eq:refined}}:
\begin{equation}
\mathbf{H}_{\text{refined}} = \text{TransformerEncoder}(\mathbf{H}_{\text{cat}}),
\label{eq:refined}
\end{equation}
which produces a refined representation \(\mathbf{H}_{\text{refined}}\). This encoder follows the multi-head self-attention framework introduced by \cite{vaswani2017attention}, allowing it to capture higher-order interactions between tokens. Although the cross-attention mechanism fuses graph-based and transformer-based contexts, an additional Transformer pass enables the model to learn nuanced relationships that may not be fully captured by a single GAT layer or a single cross-attention step. For instance, tokens that are closely linked in the syntactic graph might also exhibit semantic similarity, and this extra encoding stage provides a unified space for both structural and semantic cues to interact. By refining the combined representation, the model ultimately benefits from a richer, multi-hop understanding of the sentence’s syntactic and semantic properties.

\subsection{Aspect Representation Extraction}

In order to focus on aspect-specific cues, we apply a specialized GAT operation over an adjacency matrix designed to prioritize tokens near the aspect term. Let \(\mathbf{A}_{\text{aspect}}\) represent this adjacency structure, which can be constructed to highlight tokens in close proximity to the aspect location or to reuse \(\mathbf{A}_{\text{syntax}}\) restricted to aspect nodes. Unlike the earlier GAT steps that operate on more global embeddings, this aspect-focused GAT targets the region relevant to the aspect. Suppose we take the refined embeddings \(\mathbf{H}_{\text{refined}}\) as input (though one may also use the original \(\mathbf{H}\), depending on the design). Formally, we compute (\textcolor{blue}{Equation~\ref{eq:aspect_gat}}):
\begin{equation}
\mathbf{H}_{\text{aspect}} 
= \text{GATConv}\bigl(\mathbf{H}_{\text{refined}}, \mathbf{A}_{\text{aspect}}\bigr),
\label{eq:aspect_gat}
\end{equation}
where \(\mathbf{H}_{\text{refined}}\) provides token representations enriched by both transformer and cross-attention steps, and \(\mathbf{A}_{\text{aspect}}\) restricts the attention mechanism to aspect-centric dependencies.

Once this GAT layer is applied, each token's embedding in \(\mathbf{H}_{\text{aspect}}\) incorporates information specifically relevant to the aspect. We then extract a single vector to represent the entire aspect by pooling over the indices spanned by the aspect term (\textcolor{blue}{Equation~\ref{eq:aspect_pool}}):
\begin{equation}
\mathbf{z}_{\text{aspect}} 
= \frac{1}{|M|} \sum_{j=1}^{M} \mathbf{H}_{\text{aspect}}[j],
\label{eq:aspect_pool}
\end{equation}
where \(|M|\) denotes the number of tokens forming the aspect, and \([1,\ldots,M]\) are the corresponding indices in \(\mathbf{H}_{\text{aspect}}\). This pooling step yields a compact embedding \(\mathbf{z}_{\text{aspect}}\) that concentrates on the region of the sentence determining sentiment polarity. By focusing on aspect-specific dependencies, the model can disentangle general context from cues most pertinent to the aspect, ultimately enhancing its ability to capture fine-grained sentiment relationships in ABSA.

\begin{table}[!ht]
\centering
\caption{\small Data distribution for the four domain-specific ABSA datasets in Bangla.}
\label{dataset_stats}
\begin{tabular}{l|r|r|r}
\hline
\multirow{2}{*}{\textbf{Dataset}} & \multicolumn{3}{c}{\textbf{\# Samples}} \\
\cline{2-4}
& \textbf{Train} & \textbf{Validation} & \textbf{Test} \\
\hline
Car         & 1700 & 213 & 213 \\
Mobile      & 1486 & 186 & 186 \\
Movie       & 1184 & 148 & 148 \\
Restaurant  & 1207 & 151 & 151 \\
\hline
\end{tabular}
\end{table}

\begin{table}[!ht]
    \centering
    \caption{\small Statistics of SemEval 2014 Task 4 datasets (Laptop and Restaurant) with sentiment distribution.}
    \label{semeval_dataset}
    \begin{tabular}{l l r r r}
        \toprule
        \textbf{Dataset} & \textbf{Split} & \textbf{Positive} & \textbf{Neutral} & \textbf{Negative} \\
        \midrule
        \multirow{2}{*}{Laptop} 
            & Train & 976 & 455 & 851 \\
            & Test  & 337 & 167 & 128 \\
        \midrule
        \multirow{2}{*}{Restaurant} 
            & Train & 2164 & 637 & 807 \\
            & Test  & 727 & 196 & 196 \\
        \bottomrule
    \end{tabular}
\end{table}

\begin{table}[!ht]
\centering
\small
\caption{\small Hyperparameter Settings and Additional Techniques}
\label{tab:hyperparams}
\begin{tabular}{l c c}
\toprule
\textbf{Hyperparameter} & \textbf{Candidate Values} & \textbf{Best} \\
\midrule
Batch Size        & \{1, 2, 3\}                     & 1                    \\
Maximum Length    & \{80, 100, 128\}                & 128                  \\
Epochs            & \{10, 20, 30\}                  & 20                   \\
Learning Rate     & \{0.01, 0.001, 1e-5, 2e-5\}      & 2e-5                 \\
GATConv Layers    & \{1, 2, 3, 4, 5, 6, 7\}          & 1                    \\
Optimizer         & \{Adam, AdamW\}                 & AdamW  \\
Class Weights     & \{None, Weighted\}              & Weighted             \\
Hidden Dimension  & \{256, 512, 768\}               & 768                  \\
\bottomrule
\end{tabular}

\end{table}

\begin{table*}[!ht]
    \centering
    \small
    \caption{\small Comparative performance of three embedding-based approaches (mBERT, RoBERTa, BanglaBERT) on four domain-specific Bengali ABSA datasets. Acc and F1 are reported in percentage. The best results in each column are highlighted in bold.}
    \renewcommand{\arraystretch}{1.2}
    \begin{tabular}{l r r r r r r r r}
        \toprule
        \textbf{Model} & \multicolumn{2}{c}{\textbf{Car (\%)}} & \multicolumn{2}{c}{\textbf{Mobile (\%)}} & \multicolumn{2}{c}{\textbf{Movie (\%)}} & \multicolumn{2}{c}{\textbf{Restaurant (\%)}} \\
        \cmidrule(lr){2-3} \cmidrule(lr){4-5} \cmidrule(lr){6-7} \cmidrule(lr){8-9}
                       & \textbf{Acc$\uparrow$}   & \textbf{ F1$\uparrow$} & \textbf{Acc$\uparrow$}   & \textbf{ F1$\uparrow$} & \textbf{Acc$\uparrow$}   & \textbf{ F1$\uparrow$} & \textbf{Acc$\uparrow$}   & \textbf{ F1$\uparrow$} \\
        \midrule
         BanglaBERT & 61.50 & 61.68 & 69.35 & 69.07 & 64.19 & 65.41 & 54.97 & 54.90 \\ 
        mBERT      & \textbf{70.89} & \textbf{68.37} & 68.28 & 66.04 & \textbf{86.49} & \textbf{86.32} & 76.16 & 76.16 \\
        RoBERTa    & 68.54 & 67.29 & \textbf{79.57} & \textbf{79.44} & 70.95 & 71.95 & \textbf{88.74} & \textbf{88.69} \\

        \bottomrule
    \end{tabular}

    \label{results}
\end{table*}

\begin{table*}[!ht]
    \centering
    \small
    \caption{\small Performance comparison of our proposed model with baseline models on multiple datasets. The best results are highlighted in bold, and the second-best results are underlined.}
    \renewcommand{\arraystretch}{1.2}
    \begin{tabular}{l r r r r r r r r}
        \toprule
        \textbf{Model} & \multicolumn{2}{c}{\textbf{Car (\%)}} & \multicolumn{2}{c}{\textbf{Mobile (\%)}} & \multicolumn{2}{c}{\textbf{Movie (\%)}} & \multicolumn{2}{c}{\textbf{Restaurant (\%)}} \\
        \cmidrule(lr){2-3} \cmidrule(lr){4-5} \cmidrule(lr){6-7} \cmidrule(lr){8-9}
        & \textbf{Acc$\uparrow$} & \textbf{ F1$\uparrow$} & \textbf{Acc$\uparrow$} & \textbf{ F1$\uparrow$} & \textbf{Acc$\uparrow$} & \textbf{ F1$\uparrow$} & \textbf{Acc$\uparrow$} & \textbf{ F1$\uparrow$} \\
        \midrule
        Roberta-base (Zero-shot)  & 50.75 & 46.51 & 52.53 & 48.59 & 63.24 & 59.07 & 57.65 & 51.68 \\
        Roberta-large (Zero-shot) & 54.52 & 57.91 & 68.82 & 67.46 & 66.08 & 65.58 & 71.82 & 70.68 \\
        Roberta                   & 56.34 & 50.41 & 68.28 & 68.16 & \underline{74.32} & \underline{74.46} & \underline{80.13} & \underline{79.91} \\
        BanglaBERT                & 56.81 & 41.16 & 48.39 & 31.56 & 68.24 & 55.36 & 54.97 & 38.99 \\
        GCN                       & 64.79 & 64.63 & 67.74 & 67.75 & 73.67 & 74.34 & 68.21 & 68.25 \\
        Transformers              & 66.20 & 65.79 & \underline{76.34} & \underline{76.27} & 71.62 & 71.45 & 74.83 & 74.47 \\
        mBERT                     & 67.14 & 65.76 & 68.82 & 68.42 & 70.95 & 70.26 & 71.52 & 71.45 \\
        BiLSTM                    & 68.08 & \underline{68.22} & 70.97 & 70.44 & 68.24 & 69.17 & 72.19 & 72.19 \\
        BiGRU                     & 68.08 & 68.22 & 72.04 & 72.05 & 71.62 & 71.06 & 75.50 & 75.44 \\
        Dual-GCN                  & \underline{69.95} & 67.73 & 70.43 & 70.09 & 72.97 & 73.65 & 70.20 & 70.18 \\ \midrule
        \textbf{CrosGrpsABS}   & \textbf{70.89} & \textbf{68.37} & \textbf{79.57} & \textbf{79.44} & \textbf{86.49} & \textbf{86.32} & \textbf{88.74} & \textbf{88.69} \\
        \bottomrule
    \end{tabular}
    \label{model_comparison}
\end{table*}

\subsection{Highway Gating Mechanism}

To integrate the refined representation \(\mathbf{H}_{\text{refined}}\) with the aspect vector \(\mathbf{z}_{\text{aspect}}\) in a controlled manner, we employ a Highway Network gating function \cite{srivastava2015highway}. First, we condense the refined token representations into a single vector by mean pooling:
\begin{equation}
\bar{\mathbf{h}} 
= \frac{1}{T} \sum_{i=1}^{T} \mathbf{H}_{\text{refined}}[i],
\label{eq:mean_pool}
\end{equation}
where \(\bar{\mathbf{h}}\) aggregates the overall context across all tokens. Next, we form a composite representation \(\mathbf{u}\) by concatenating \(\bar{\mathbf{h}}\) with the aspect embedding \(\mathbf{z}_{\text{aspect}}\):
\begin{equation}
\mathbf{u} 
= \bigl[\bar{\mathbf{h}} \,\|\, \mathbf{z}_{\text{aspect}}\bigr] \in \mathbb{R}^{2d},
\label{eq:concat_u}
\end{equation}
which ensures that both global contextual cues and aspect-specific features are available to the gating mechanism.

\begin{table}[!ht]
\centering
\small
\caption{\small Comparison of Different Models on ABSA Tasks (SemEval 2014 Task 4 datasets: Restaurant and Laptop)}
\label{multirow_example}
\renewcommand{\arraystretch}{1.2}
\begin{tabular}{l r r r r}
\toprule
\multirow{2}{*}{\textbf{Models}} 
 & \multicolumn{2}{c}{\textbf{Restaurant (\%)}} 
 & \multicolumn{2}{c}{\textbf{Laptop (\%)}} \\
\cmidrule(lr){2-3}\cmidrule(lr){4-5}
 & \textbf{Acc$\uparrow$} & \textbf{F1$\uparrow$} 
 & \textbf{Acc$\uparrow$} & \textbf{F1$\uparrow$} \\
\midrule
BERT \cite{devlin2019bert}     & 85.97 & 82.79 & 79.10 & 76.00 \\
R-GAT+BERT \cite{wang2020relational}   & 86.60 & 83.00 & 80.30 & 77.47 \\
DGEDT+BERT \cite{tang2020dependency}  & 82.00 & 80.00 & 79.80 & 80.79 \\
BART-GCN \cite{xiao2021bert4gcn} & 84.75 & 77.11 & 79.19 & 76.00 \\ 
T-GCN+BERT \cite{yin2024textgt} & 85.00 & 80.00 & 80.00 & 77.00 \\ 
DualGCN+BERT \cite{li2021dual} & 87.13 & 82.16 & 81.80 & 76.00 \\
SSEGCN+BERT \cite{zhang2022ssegcn} & 85.16 & 81.00 & 79.00 & 76.00 \\
AG-VSR+BERT \cite{yin2024textgt} & \textbf{89.96} & \underline{85.00} & 80.00 & 78.00 \\
TextGT+BERT \cite{yin2024textgt} & 87.31 & 82.27 & 81.33 & 78.71 \\
DAGCN+BERT \cite{wang2024dagcn} & 88.03 & 82.64 & \underline{82.59} & \underline{79.40} \\
\midrule
\textbf{CrosGrpsABS} & \underline{87.88} & \textbf{85.93} & \textbf{86.43} & \textbf{80.46} \\ 
\bottomrule
\end{tabular}
\end{table}

\begin{table*}[!ht]
    \centering
    \small
    \caption{\small Ablation study results for our proposed model across multiple datasets. The best results are highlighted in bold.}
    \renewcommand{\arraystretch}{1.2}
    \begin{tabular}{l r r r r r r r r}
        \toprule
        \textbf{Settings} & \multicolumn{2}{c}{\textbf{Car (\%)}} & \multicolumn{2}{c}{\textbf{Mobile (\%)}} & \multicolumn{2}{c}{\textbf{Movie (\%)}} & \multicolumn{2}{c}{\textbf{Restaurant (\%)}} \\
        \cmidrule(lr){2-3} \cmidrule(lr){4-5} \cmidrule(lr){6-7} \cmidrule(lr){8-9}
        & Acc & F1 & Acc &  F1 & Acc &  F1 & Acc &  F1 \\
        \midrule
        No Syntax Graph  & 58.70  & 53.24  & 70.77  & 70.75  & 69.59  & 70.61  & 79.47  & 79.13  \\
        No Semantic Graph  & 64.78  & 59.25  & 47.18  & 30.25  & 70.27  & 70.58  & 83.44  & 83.45  \\
        No Graph Branches & 67.87  & 67.75  & 74.36  & 74.29  & 71.62  & 72.25  & 76.16  & 76.20  \\
        No Cross-Attention  & 68.26  & 68.10  & 69.23  & 69.25  & 78.38  & 78.50  & 78.15  & 78.18  \\
        No Transformer  & 61.30  & 56.77  & 75.38  & 74.97  & 77.70  & 78.34  & 82.78  & 82.78  \\
        No Highway Gate  & 63.91  & 59.69  & 74.87  & 74.42  & 74.32  & 74.46  & 80.79  & 80.78  \\
        No Aspect Embedding  & 69.13  & 67.10  & 73.85  & 72.67  & 78.38  & 75.11  & 81.77  & 81.76  \\
        Fixed Adjacency  & 64.35  & 58.90  & 70.26  & 70.22  & 75.68  & 76.35  & 79.66  & 77.63  \\ \midrule
        \textbf{CrosGrpsABS Base}   & \textbf{70.89} & \textbf{68.37} & \textbf{79.57} & \textbf{79.44} & \textbf{86.49} & \textbf{86.32} & \textbf{88.74} & \textbf{88.69} \\
        \bottomrule
    \end{tabular}
    
    \label{ablation_study}
\end{table*}

We then define two functions acting on \(\mathbf{u}\). The first, \(\mathbf{H}(\cdot)\), is a feed-forward transformation (e.g., a ReLU-activated linear layer) that generates a candidate set of new features. The second, \(\mathbf{T}(\cdot)\), is a gating function that produces a sigmoid output indicating how much of these new features should be integrated into the final representation \cite{he2016deep}. Specifically, we compute
\begin{equation}
\mathbf{z}_H 
= \mathbf{T}(\mathbf{u}) \,\odot\, \mathbf{H}(\mathbf{u}) 
  \;+\; \bigl(\mathbf{1} - \mathbf{T}(\mathbf{u})\bigr) \,\odot\, \mathbf{u},
\label{eq:highway_gate}
\end{equation}
where 
\(\mathbf{T}(\mathbf{u}) = \sigma\bigl(\mathbf{W}_T \mathbf{u} + \mathbf{b}_T\bigr)\).
In \textcolor{blue}{Equation~\eqref{eq:highway_gate}}, the gating vector \(\mathbf{T}(\mathbf{u})\) determines the balance between newly transformed information \(\mathbf{H}(\mathbf{u})\) and the original input \(\mathbf{u}\). As a result, tokens closely related to the aspect can be emphasized, while broader contextual cues remain accessible. This design also helps mitigate vanishing gradients by allowing unimpeded flow of input features (i.e., skip connections), making the gating mechanism robust in deeper networks \cite{srivastava2015highway,he2016deep}. Consequently, the final output \(\mathbf{z}_H\) represents a balanced fusion of aspect-specific cues and the preserved global context, thereby enhancing the model’s ability to capture nuanced sentiment relationships.

\subsection{Classification Layer}
In the final stage, the fused representation $\mathbf{z}_H$ is mapped to sentiment class probabilities. A linear projection transforms $\mathbf{z}_H$ into a vector of logits corresponding to the $C$ sentiment classes (e.g., positive, negative, neutral). The predicted label $\hat{y}$ is then obtained by selecting the index of the highest logit. During training, cross-entropy loss is used to measure the alignment between the predicted probabilities and the true class, thereby fine-tuning all model components including gating, cross-attention, and graph-based modules to better distinguish among sentiment polarities.

\section{Experimental Setup}\label{experimental}
\subsection{Datasets}
We conduct experiments on four public benchmark datasets for aspect-based sentiment analysis (ABSA) in Bangla, introduced by Hasan et al.~\cite{hasan2024banglaabsa}, as well as on English datasets from the widely used SemEval 2014 Task 4 benchmark~\cite{pontiki2014semeval}.

Our Bangla datasets consist of {Car}, {Mobile Phone}, {Movie}, and {Restaurant}, each provided in a dedicated Excel file ({Car ABSA}, {Mobile phone ABSA}, {Movie ABSA}, and {Restaurant ABSA}, respectively) with three columns: {Id}, {Comment}, and {Aspect category, Sentiment Polarity}. These datasets were compiled from platforms such as Facebook, YouTube, and various online portals. Specifically, {Car} has 1149 samples, {Mobile Phone} has 975, {Movie} contains 800, and {Restaurant} has 801, collectively offering a comprehensive resource for ABSA in Bangla. \textcolor{blue}{Table \ref{dataset_stats}} presents our data distribution for the four domain-specific ABSA datasets in Bangla.

In addition, we evaluate on the SemEval 2014 Task 4 English datasets, which include the {Laptop} and {Restaurant} domains~\cite{pontiki2014semeval}. These datasets have been extensively used in previous ABSA research and serve as a standard benchmark for comparing model performance across languages. Each dataset contains sentences annotated with aspect terms and associated sentiment polarities (positive, negative, or neutral), ensuring a fair comparison with state-of-the-art ABSA approaches. \textcolor{blue}{Table \ref{semeval_dataset}} presents statistics of SemEval 2014 Task 4 datasets (Laptop and Restaurant) with sentiment distribution.

\subsection{Data Preparation}
In preparing the data, we first perform a series of cleaning steps to remove noise, such as correcting typographical errors, removing irrelevant symbols or URLs, and normalizing character encodings. This ensures a consistent and high-quality corpus for subsequent processing. Next, we employ a pre-trained tokenizer to segment the text into subword units, leveraging the tokenizer’s learned vocabulary to capture linguistic nuances and reduce out-of-vocabulary issues. By combining thorough data cleaning with an advanced, pre-trained tokenizer \cite{liu2019roberta}, we create a reliable foundation for downstream tasks while minimizing potential errors and preserving meaningful textual information.

\subsection{Baselines}
We compare our proposed \textbf{CrosGrpsABS} with several strong baseline models to evaluate its effectiveness. The baselines include Transformer-based models employing pre-trained embeddings: RoBERTa-base and RoBERTa-large evaluated under zero-shot conditions~\cite{liu2019roberta}, fine-tuned RoBERTa~\cite{liu2019roberta}, multilingual BERT (mBERT)~\cite{devlin2019bert}, and BanglaBERT~\cite{banglabert}, a transformer explicitly pretrained on Bangla text. Additionally, we incorporate neural architectures such as Graph Convolutional Networks (GCN), Dual-GCN, Bidirectional LSTM (BiLSTM), and Bidirectional GRU (BiGRU), all enhanced with Transformer-based embeddings. These models provide comprehensive comparative benchmarks, reflecting state-of-the-art approaches commonly employed in ABSA tasks.
\paragraph{Evaluation on SemEval 2014 Task 4} We further evaluate the performance of our proposed model, \textbf{CrosGrpsABS}, on the widely-used SemEval 2014 Task 4 benchmark datasets, specifically the Laptop and Restaurant domains. To provide a rigorous comparison, we include several state-of-the-art baselines previously established in ABSA literature, including BERT~\cite{devlin2019bert}, R-GAT+BERT~\cite{wang2020relational}, DGEDT+BERT~\cite{tang2020dependency}, BART-GCN \cite{xiao2021bert4gcn}, T-GCN+BERT \cite{yin2024textgt}, DualGCN+BERT \cite{li2021dual}, SSEG-CN+BERT \cite{zhang2022ssegcn}, AG-VSR+BERT \cite{yin2024textgt}, TextGT+-BERT \cite{yin2024textgt}, and DAGCN+BERT \cite{wang2024dagcn}. These strong baselines enable a comprehensive assessment of our model's capability to generalize across different languages and datasets.
\begin{figure}[htbp]
  \centering
  \includegraphics[scale=0.9]{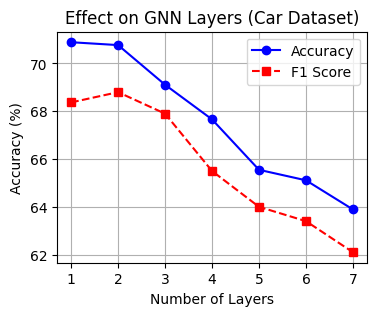}
    \caption{\small Effect of GNN Layers on Performance (Car Dataset). The chart compares Accuracy and F1 Score across different numbers of GNN layers, ranging from 1 to 7.}
  \label{gnn_layers_car}
\end{figure}

\subsection{Ablation Study}
To systematically evaluate the contribution of each component within our proposed model, \textbf{CrosGrpsABS}, we conduct extensive ablation experiments under various settings. Specifically, we investigate the impact of removing key components, including the syntactic graph (No Syntax Graph), semantic graph (No Semantic Graph), both graph branches entirely (No Graph Branches), the cross-attention mechanism (No Cross-Attention), Transformer embeddings (No Transformer), the highway gating mechanism (No Highway Gate), aspect embeddings (No Aspect Embedding), and utilizing a fixed adjacency matrix (Fixed Adjacency). These analyses provide deeper insights into the effectiveness of individual modules and their interactions within our proposed architecture.
\subsection{Computational Resources}
All experiments were conducted using Python 3.12.9 and PyTorch 2.4, utilizing an NVIDIA GeForce RTX 2060 GPU with 16 GB of RAM. This hardware and software configuration ensured efficient training and evaluation of our models, allowing reproducible and reliable comparisons across various baseline methods and proposed architectures.
\subsection{Evaluation Metrics}
To assess the performance of our proposed model and baseline methods, we utilize two widely adopted evaluation metrics: Test Accuracy (Acc) and Micro-averaged F1 score (F1). Accuracy provides a straightforward measure of overall correctness, while the Micro F1 score offers a balanced evaluation by considering both precision and recall across all classes, particularly beneficial for datasets with class imbalance. These metrics collectively enable comprehensive and reliable evaluation of our aspect-based sentiment classification approach.

\subsection{Hyperparameter Settings}
\noindent
\textcolor{blue}{Table~\ref{tab:hyperparams}} summarizes the hyperparameter settings and additional techniques used in our experiments. We evaluated multiple batch sizes and sequence lengths, identifying a batch size of 1 and a maximum sequence length of 128 as optimal. Training for 20 epochs at a learning rate of \(2\times10^{-5}\) balanced convergence speed and stability. We chose a single GATConv layer for simplicity and efficiency, while AdamW (with a weight decay of \(1\times10^{-2}\)) served as our optimizer. To address potential overfitting, we used early stopping callbacks, and we applied class weights to mitigate imbalance. Finally, a hidden dimension of 768 provided ample capacity for capturing nuanced linguistic representations.

\section{Results and Discussions}\label{resullts}
\textcolor{blue}{Table~\ref{results}} presents the comparative performance of our proposed \textbf{CrosGrpsABS} model using three different embedding techniques: BanglaBERT, mBERT~\cite{devlin2019bert}, and RoBERTa~\cite{liu2019roberta}, evaluated across four domain-specific Bengali ABSA datasets: Car, Mobile, Movie, and Restaurant. The results are reported in terms of accuracy (Acc) and micro F1 scores (F1). Among these embedding techniques, RoBERTa-based embeddings achieve the highest performance in most domains, notably in Car, Movie, and Restaurant datasets, with a peak accuracy of 88.74\% and an F1 score of 88.69\% on the Restaurant dataset. The mBERT embeddings yield competitive results, particularly in the Mobile domain, achieving the highest accuracy (86.49\%) and micro F1 (86.32\%). Conversely, the BanglaBERT embeddings lead to relatively lower performance, suggesting the advantage of utilizing multilingual transformer-based embeddings like RoBERTa and mBERT within the \textbf{CrosGrpsABS} framework.

\begin{figure}[htbp]
  \centering
  \includegraphics[scale=0.66]{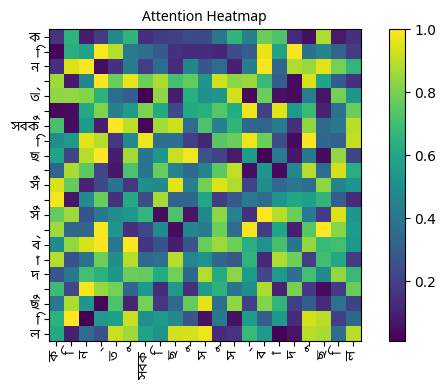}
  \caption{\small Attention heatmap illustrating the token-to-token attention weights for a sample Bengali sentence. Higher-intensity regions (yellow) indicate stronger attention signals between corresponding tokens.}
  \label{attention}
\end{figure}

\subsection{Baseline Comparison}
\textcolor{blue}{Table~\ref{model_comparison}} presents the comparative performance of our proposed \textbf{CrosGrpsABS} model against multiple baseline models on four domain-specific Bengali ABSA datasets: Car, Mobile, Movie, and Restaurant. The baseline methods include transformer-based embeddings (RoBERTa-base and RoBERTa-large evaluated under zero-shot settings, RoBERTa, BanglaBERT~\cite{banglabert}, mBERT~\cite{devlin2019bert, liu2019roberta}), as well as graph-based models (GCN, Dual-GCN), sequential neural networks (BiLSTM, BiGRU), and standard Transformers architecture. Our \textbf{CrosGrpsABS} model consistently achieves the highest performance across all domains, significantly outperforming all baseline models. Specifically, \textbf{CrosGrpsABS} attains its best results in the Restaurant dataset, achieving an accuracy of 88.74\% and a micro F1 of 88.69\%. Among the baselines, fine-tuned RoBERTa shows robust performance and emerges as the strongest baseline, consistently ranking second-best, notably with 80.13\% accuracy on Restaurant data. The lower performance of zero-shot RoBERTa-base and RoBERTa-large across all domains highlights the importance of task-specific fine-tuning. Meanwhile, BanglaBERT exhibits moderate performance, emphasizing that even language-specific pretraining alone may not suffice without further adaptations. Sequential models such as BiLSTM and BiGRU, and the graph-based models (GCN, Dual-GCN) demonstrate competitive but inferior performance, validating the effectiveness of our hybrid approach that integrates transformer-based embeddings with bidirectional cross-attention on syntactic and semantic graphs.
\begin{figure*}[htbp]
  \centering
  \includegraphics[scale=0.75]{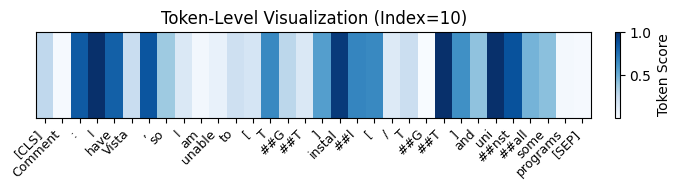}
  \includegraphics[scale=0.75]{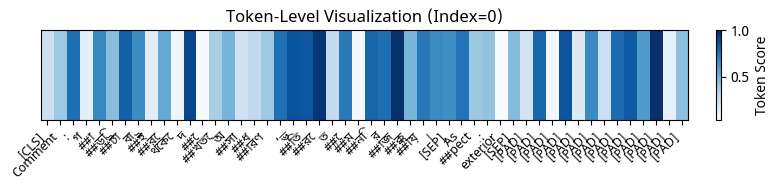}
  \caption{\small Token-level visualization illustrating the importance scores assigned by CrosGrpsABS. The top visualization is from the SemEval 2014 Task 4 (Laptop domain, English), and the bottom visualization is from our Bengali ABSA dataset. Darker shades represent higher token importance in predicting aspect-level sentiment.}
  \label{token}
\end{figure*}

\paragraph{SemEval 2014 Task 4 results: }
\textcolor{blue}{Table~\ref{multirow_example}} provides a comprehensive comparison of our proposed \textbf{CrosGrpsABS} model against state-of-the-art baseline methods on the SemEval 2014 Task 4 datasets (Restaurant and Laptop). The baselines include recent advanced ABSA methods such as BERT, R-GAT+BERT, DGEDT+BERT, BART-GCN, T-GCN+BERT, DualGCN+BERT, SSEG-CN+BERT, AG-VSR+BERT, Tex-tGT+BERT, and DA-GCN+BERT. In the Restaurant dataset, \textbf{CrosGrpsABS} achieves highly competitive results with an accuracy of 87.88\% and micro F1 score of 87.93\%, slightly below AG-VSR+BERT, which attains the highest accuracy at 89.96\%. On the Laptop dataset, \textbf{CrosGrpsABS} clearly demonstrates superior performance by reaching the highest accuracy (86.43\%) and micro F1 score (86.46\%), surpassing all other baseline methods. These results validate the effectiveness and robustness of \textbf{CrosGrpsABS} across different datasets and domains, emphasizing its capability to compete with state-of-the-art ABSA approaches.

    
\begin{table*}[t]
  \centering
  \caption{\small Examples from the Bengali ABSA Dataset with True Labels and Model Predictions.}
  \begin{tabular}{@{}c@{}} 
    \includegraphics[width=\linewidth]{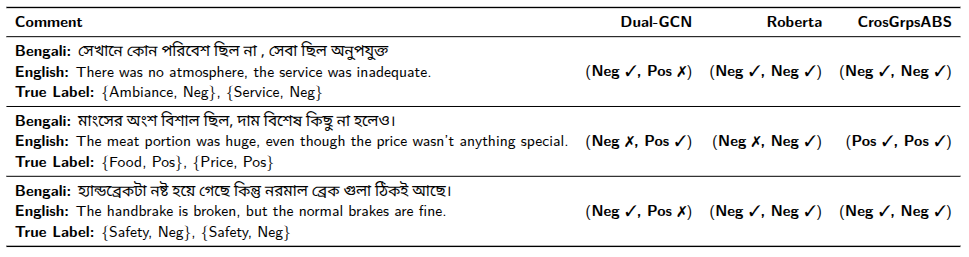}
  \end{tabular}
  \label{case}
\end{table*}

\subsection{Ablation Study}
\textcolor{blue}{Table~\ref{ablation_study}} presents an ablation study conducted on our proposed \textbf{CrosGrpsABS} model, evaluating the contributions of different components across four Bengali ABSA datasets: Car, Mobile, Movie, and Restaurant. Specifically, we examine the effects of removing individual modules, including the syntactic graph, semantic graph, both graph branches, cross-attention, transformer embeddings, highway gating mechanism, aspect embeddings, and using a fixed adjacency matrix. The results demonstrate that the complete \textbf{CrosGrpsABS} model consistently outperforms all ablated variants, emphasizing the critical role of each component. Notably, removing the semantic graph significantly degrades performance, especially in the Mobile domain, dropping accuracy from 79.57\% to 47.18\%, which highlights the importance of semantic information. Excluding transformer embeddings also leads to notable performance reductions across all datasets, indicating transformers' key role in encoding contextual information. Similarly, eliminating the syntactic graph results in decreased accuracy and F1 scores, particularly evident in the Car dataset, reducing accuracy from 70.89\% to 58.70\%. These findings collectively confirm the effectiveness and necessity of integrating both syntactic and semantic information with transformer embeddings and cross-attention mechanisms to achieve optimal performance.

\subsection{Effect on GNN layers}
\textcolor{blue}{Figure~\ref{gnn_layers_car}} shows how varying the number of GNN layers (from 1 to 7) impacts both Accuracy and F1 Score on the Car dataset. Notably, models with shallower architectures (one or two layers) attain the highest performance, suggesting that adding more layers can lead to overfitting or oversmoothing, where critical signals become diluted in deeper networks. While intermediate depths still yield competitive metrics, the overall trend highlights a gradual decline in performance as the layer count increases. This indicates the importance of selecting an optimal GNN depth to balance representational capacity against potential training instabilities, thereby avoiding unnecessary complexity that can undermine predictive accuracy and generalization.

\subsection{Visualization}
\textcolor{blue}{Figure~\ref{attention}} illustrates an attention heatmap generated by our \textbf{CrosGrpsABS} model, demonstrating token-to-token attention weights for a sample Bengali sentence. The heatmap visualizes attention intensity between pairs of tokens, with higher-intensity regions (indicated by yellow) representing stronger attention signals. This visualization highlights how \textbf{CrosGrpsABS} effectively captures the relationships between different words within the sentence, providing insights into how syntactic and semantic information is integrated through cross-attention mechanisms. Such attention visualization further aids interpretability by allowing analysis of the model's reasoning process when identifying sentiment-bearing aspects in Bengali texts.

\textcolor{blue}{Figure~\ref{token}} presents token-level visualizations of attention scores produced by our \textbf{CrosGrpsABS} model for two distinct ABSA datasets: the SemEval 2014 Task 4 (Laptop domain, top) and our Bengali ABSA dataset (bottom). In each visualization, darker shades of blue correspond to higher token importance, reflecting stronger contributions of specific tokens to the sentiment classification decision. In the SemEval dataset example, the model effectively emphasizes tokens such as \textit{``vista''}, \textit{``unable''}, and \textit{``install''}, highlighting its capability to detect negative sentiment expressions associated with a particular aspect (e.g., software compatibility). In the Bengali dataset example, the visualization similarly highlights tokens critical to sentiment analysis, illustrating how the \textbf{CrosGrpsABS} model successfully identifies important Bengali linguistic elements and context-specific expressions. The visualizations demonstrate that \textbf{CrosGrpsABS} not only performs well quantitatively but also provides qualitative interpretability by clearly indicating which tokens drive its sentiment predictions. This interpretability enhances confidence in the model's decision-making processes and facilitates deeper insights into aspect-level sentiment detection across languages and domains.

\subsection{Case Study}
\noindent\textcolor{blue}{Table~\ref{case}} presents a detailed case study analysis, showcasing three challenging Bengali ABSA examples alongside predictions from Dual-GCN, RoBERTa, and our proposed \textbf{CrosGrpsABS} model. In the first example, both RoBERTa and \textbf{CrosGrpsABS} correctly identify negative sentiment for both \textit{Ambiance} and \textit{Service}, whereas Dual-GCN misclassifies \textit{Service} as positive. The second example involves two positive aspects (\textit{Food} and \textit{Price}); Dual-GCN and RoBERTa each misjudge at least one of these aspects, while \textbf{CrosGrpsABS} accurately predicts both. Finally, the third example contains a contrasting statement (\textit{handbrake broken} vs.\ \textit{normal brakes fine}), and although Dual-GCN misclassifies the second \textit{Safety} aspect, both RoBERTa and \textbf{CrosGrpsABS} correctly predict negative sentiment for both. These observations highlight \textbf{CrosGrpsABS}’s robustness in handling complex linguistic cues and mixed sentiment contexts within Bengali text.

\section{Conclusion}\label{conclusion}
In this work, we proposed \textbf{CrosGrpsABS}, a novel hybrid model designed specifically to enhance Aspect-Based Sentiment Analysis (ABSA) for the low-resource Bengali language. Our approach uniquely integrates transformer-based embeddings with bidirectional cross-attention mechanisms on syntactic and semantic graph representations, effectively combining local syntactic cues with global semantic contexts. Extensive experiments conducted across multiple domain-specific Bengali datasets, as well as the SemEval 2014 benchmark datasets, have demonstrated the superior performance of \textbf{CrosGrpsABS} compared to several state-of-the-art baselines. Moreover, qualitative analyses, such as attention visualizations and token-level importance scores, have provided valuable insights into the interpretability and reasoning mechanisms of our proposed framework.

However, despite the overall effectiveness of \textbf{CrosGrpsABS}, certain limitations remain evident. Specifically, the model sometimes struggles with sentences expressing mixed sentiments, implicit emotional expressions, or contexts involving contrasting or comparative aspects. Such cases often require deeper semantic comprehension and nuanced reasoning beyond the model's current capabilities. Another limitation pertains to the model's reliance on accurate syntactic dependency parsing, which can be challenging for languages with limited linguistic resources, potentially affecting the model’s performance in less structured scenarios. Potential improvements include developing advanced contrastive and context-aware learning techniques to better handle sentences with implicit or mixed sentiments. Additionally, exploring multilingual or cross-lingual transfer learning approaches could further improve the robustness and generalizability of ABSA methods for low-resource languages. Future research could also investigate more sophisticated graph-based techniques, such as dynamic adjacency mechanisms or adaptive graph construction, to further refine the modeling of syntactic and semantic dependencies. Finally, expanding this methodology to other low-resource languages and diverse domains could significantly enhance its practical applicability, thereby advancing the broader field of ABSA.



\section*{Conflits of Interest}
The authors declare that they have no conflict of interest.


\printcredits



\bibliography{cas-refs}

\end{document}